# Use of semantic technologies for the development of a dynamic trajectories generator in a Semantic Chemistry eLearning platform

Richard Huber, Kirsten Hantelmann

FIZ Chemie Berlin
[huber | hantelmann] AT fiz-chemie.de

Alexandru Todor, Sebastian Krebs, Ralf Heese, Adrian Paschke

Freie Universität Berlin
[todor | krebs | heese | paschke] AT inf.fu-berlin.de

**ABSTRACT** ChemgaPedia is a multimedia, webbased eLearning service platform that currently contains about 18.000 pages organized in 1.700 chapters covering the complete bachelor studies in chemistry and related topics of chemistry, pharmacy, and life sciences. The eLearning encyclopedia contains some 25.000 media objects and the eLearning platform provides services such as virtual and remote labs for experiments. With up to 350.000 users per month the platform is the most frequently used scientific educational service in the German spoken Internet. In this demo we show the benefit of mapping the static eLearning contents of ChemgaPedia to a Linked Data representation for Semantic Chemistry which allows for generating dynamic eLearning paths tailored to the semantic profiles of the users.

## INTRODUCTION

Learning is a very trying and individual process for any single user of an eLearning environment. Especially unskilled learners require either a human teacher/ tutor or a very intelligent, personalized artificial intelligence of the eLearning environment to be guided by an individual learning procedure and learning path. Also with more experienced learners the studying process of an individual can be optimized by personalized didactics, dynamic assembly of pages and units and in a strong way by an appropriate and distinct teaching path.

To distinguish the results of personalized, individual didactics, composition of page content and path generation from traditional eLearning content and hardcoded paths we henceforth call the dynamically generated, individual teaching paths "Dynamic eLearning Trajectories" (DeLTs).

To increase the acceptance of an educational platform this device should be able to adjust to individual skills, experiences, performance, and the personal motivation. Personalization concepts for eLearning environments can be performed best if there are large sets of user data available. The generation of user data based on traditional platform techniques requires a registration process with precise monitoring of knowledge, personal background, level of experience, performance, and additional personal skills. However, this monitoring process may cause rejection of users from certain user groups like business workers who might be afraid of surveillance by colleagues or by their company's management. Any registration process poses a barrier for free participation and decreases the amount of users of a platform. Lower numbers of users weaken the market position of the platform.

For a freely available eLearning platform there are only two ways to build up a DeLT system: Analyze session information or offer value added service in turn for registration.

In the first approach the system compares background knowledge with information of a single user which has been collected over a short period of time during his/her current session. As a result of this semantic analysis the system might recommend paths and try to deliver content aggregation in a semi-personalized way.

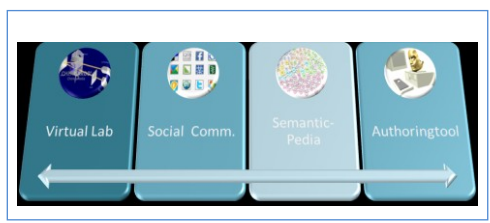

Considering the other method the original and freely available eLearning platform would be extended by value added services (cf. Fig. 1) requiring users to register if they want to benefit from them. This set of user data can then be used to develop recommendations not only for registered users but for all users of the platform. A lot of research will be necessary to find out, if this transfer of concepts and recommendations from certain user groups to all users of the eLearning environment is valid and if there are optimization procedures available to improve the transfer quality.

Fig. 1: Possible value added services to gain user data

## STATE OF THE ART

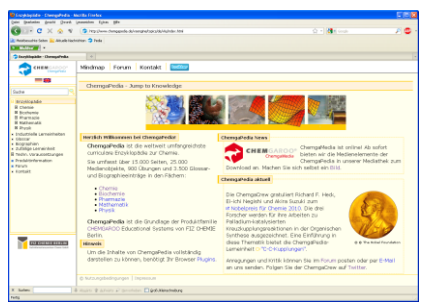

The well-established eLearning platform ChemgaPedia (http://www.chemgapedia.de) contains ca. 18.000 web pages organized in about 1.700 chapters. Fig. 2 presents a screenshot of its current version. The educational content is designed for higher education in the subjects chemistry, pharmacy, biochemistry, and related sciences. The target groups are teachers, students, experienced pupils, and vocational trainees.

Due to the high quality content the platform is up to now the most used educational environment in the German spoken chemical and pharmaceutical community. Up to 350.000 students, teachers, and trainees rely on the continuing availability of platform and content. The content within the platform consists of text and multimedia elements, such as animations, applications and videos. Teaching paths and the provision of pages and units are statically created by domain experts. The authors are responsible for producing content and relating chapters and pages.

Fig. 2: Screenshot of the current ChemgaPedia version

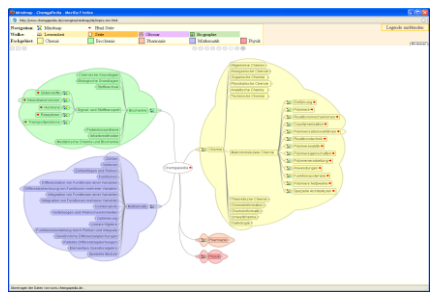

Fig. 3: Graphical representation of the hardcoded teaching path

The content is represented in XML format conforming to a schema specialized to represent eLearning content. All pages and media elements are stored in an XML-based database. A processing pipeline consisting mainly of XSLT transformations is used to convert the content to static HTML pages.

Up to now there is no method to generate dynamic paths or to assemble personalized content pages on demand.

## STRATEGY

The aim of the cooperation between FIZ CHEMIE as the provider of ChemgaPedia and the Corporate Semantic Web expert group providing know-how about semantic technologies and applications is the development of an ontology-based copy of the current ChemgaPedia platform.

For this purpose, we are currently developing two ontologies: a chemistry ontology and an eLearning ontology. In contrast to existing chemistry ontologies which are built for expert use, we use the expertise of the authors of ChemgaPedia to develop an ontology that is suitable for the use in the eLearning platform. For example, it also includes concepts that are not contained in existing ontologies but are relevant for students or it contains relationships that help to generate teaching paths automatically. Nevertheless, our goals are also to reuse existing ontologies as far as possible and to link to them in the sense of linked data.

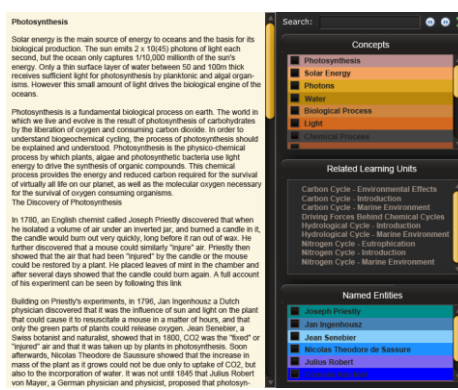

The eLearning ontology models concepts and relationships to describe the content, relationships between lectures, and metadata about a lecture, e.g., target group.

In a first step towards an ontology-based eLearning platform we currently construct the needed ontologies and develop a demonstrator illustrating the usage of background knowledge to generate teaching paths automatically. In order to represent the data in a semantic format we have built converters that parse the XML files and convert them to RDF. The converters identify chemical entities in the source content and match them to ontology concepts. Moreover, information relevant eLearning, e.g., study time, target audience, or recommended readings, is extracted as well and represented as RDF.

Fig. 4: Screenshot of the demonstration version

To reuse existing ontologies from the chemical and life-science domains being under intense development by the scientific community, we have aligned our ontological model with ontologies such as the ChEBI[1] ontology and the ChEBI and PubChem schemata from Bio2RDF[2]. This allows us to enrich the chemical data described in ChemgaPedia with chemical identifiers and classifications from these ontologies and also to improve the search functionality by adding new and more precise data. Another important step is the alignment with DBpedia Germany[3] that allows for using the elaborate SKOS based classification to enhance the learning material from ChemgaPedia with related content from DBpedia.

The demonstrator under development is based on the Open Semantic Framework (OSF). The OSF is a general purpose semantic framework composed of multiple components. The core part is StructWSF, a web service framework that provides access to a triplestore containing the eLearning data as well as allows for efficient faceted and full-text search. The display of information is handled by the conStruct component on top of Drupal, an open source content management system. Thus, it allows users to retrieve and to visualize RDF data stored in the triplestore. Since these components are designed as general purpose RDF viewers we are currently modifying their user interfaces to fulfill the needs of the eLearning domain and to allow the user to navigate through the chemistry ontology. Thus, a user is able to explore the chemistry domain and discover related learning units. A faceted search allows the user to narrow down the results.

## CONCLUSION

ChemgaPedia is a scientific educational service in the German spoken Internet, publicly available without the need to register. At the moment all its content and especially the teaching paths are statically created by domain experts. In this paper we presented the first steps towards an ontology-based eLearning platform that will be able to generate teaching paths dynamically depending on the user interests.

---

[1] http://www.ebi.ac.uk/chebi/
[2] http://bio2rdf.org/
[3] http://de.dbpedia.org/